\def\year{2021}\relax
\documentclass[letterpaper]{article} 
\usepackage{algorithm}
\usepackage{algorithmic}
\usepackage{aaai21}  
\usepackage{times}  
\usepackage{helvet} 
\usepackage[switch]{lineno}
\usepackage{courier}  
\usepackage[hyphens]{url}  
\usepackage{graphicx} 
\usepackage{natbib}  
\usepackage{caption} 
\usepackage{amsmath}
\usepackage{amssymb}
\usepackage{mathrsfs}
\usepackage{array}
\usepackage{bm}
\usepackage{multirow}
\usepackage{graphicx, subfigure}

\usepackage{xspace}

\newcommand{\E}{\mathbb{E}}

\def\KL{{\rm KL}}

\def\M{{\cal M}}

\def\tn{\tilde{n}}

\def\E{{\rm E}}

\def\tn{\tilde{n}}
\def\y{{\cal Y}}
\def\x{{\cal X}}

\makeatletter
\DeclareRobustCommand\onedot{\futurelet\@let@token\@onedot}
\def\@onedot{\ifx\@let@token.\else.\null\fi\xspace}

\makeatother

\title{Learning Cycle-Consistent Cooperative Networks via Alternating MCMC Teaching for Unsupervised Cross-Domain Translation}
\author {
    Jianwen Xie \textsuperscript{\rm 1*},
    Zilong Zheng \textsuperscript{\rm 2*},
    Xiaolin Fang \textsuperscript{\rm 3},
    Song-Chun Zhu \textsuperscript{\rm 2,4,5},
    Ying Nian Wu \textsuperscript{\rm 2} \\
}
\affiliations {
    \textsuperscript{\rm 1} Cognitive Computing Lab, Baidu Research, Bellevue, USA \\
    \textsuperscript{\rm 2} University of California, Los Angeles, USA  \\
    \textsuperscript{\rm 3} Massachusetts Institute of Technology, Cambridge, USA \\
    \textsuperscript{\rm 4} Tsinghua University, Beijing, China \\
    \textsuperscript{\rm 5} Peking University, Beijing, China\\     jianwen@ucla.edu, z.zheng@ucla.edu, xiaolinf@csail.mit.edu, sczhu@stat.ucla.edu, ywu@stat.ucla.edu\vspace{5mm}
} 

\begin{document}
\maketitle

\begin{abstract}
This paper studies the unsupervised cross-domain translation problem by proposing a generative framework, in which the probability distribution of each domain is represented by a generative cooperative network that consists of an energy-based model and a latent variable model. The use of generative cooperative network enables maximum likelihood learning of the domain model by MCMC teaching, where the energy-based model seeks to fit the data distribution of domain and distills its knowledge to the latent variable model via MCMC. Specifically, in the MCMC teaching process, the latent variable model parameterized by an encoder-decoder maps examples from the source domain to the target domain, while the energy-based model further refines the mapped results by Langevin revision such that the revised results match to the examples in the target domain in terms of the statistical properties, which are defined by the learned energy function. For the purpose of building up a correspondence between two unpaired domains, the proposed framework simultaneously learns a pair of cooperative networks with cycle consistency, accounting for a two-way translation between two domains, by alternating MCMC teaching. Experiments show that the proposed framework is useful for unsupervised image-to-image translation and unpaired image sequence translation. 
\end{abstract}


\section{Introduction}
Cross-domain translation, such as image-to-image translation, has shown its importance over the last few years on numerous computer vision and computer graphics tasks which require translating an example from one domain to another, for example, neural style transfer, photo enhancing, etc. This problem can be solved by learning a conditional generative model as a mapping from source domain to target domain in a supervised manner, when paired training examples between two domains are available. However, manually pairing up examples between two domains is costly in both time and efforts, and in some cases it is even impossible. For example, learning to translate a photo to a Van Gogh style painting requires  plenty of real scene photos paired with their corresponding paintings for training. Therefore, unsupervised cross-domain translation is considered more applicable since different domains of independent data collections are easily accessible, yet it is also regarded as a harder problem due to the lack of supervision on instance-level correspondence between different domains. This paper focuses on unsupervised cross-domain translation problem where paired training examples are not available. 

With the recent success of Generative Adversarial Networks (GANs) \cite{goodfellow2014generative,arjovsky2017wasserstein} in image generation \cite{radford2015unsupervised,denton2015deep,brock2018large}, researchers have proposed unsupervised cross-domain translation networks based on GANs and obtained compelling results \cite{zhu2017unpaired,liu2017unsupervised,huang2018multimodal}. 
For the sake of learning probability distribution, instead of maximizing the data likelihood, GANs introduce the concept of adversarial learning between a generator and a discriminator. Specifically, the generator is the desired implicit data distribution that maps the Gaussian prior on a low-dimensional latent space to the data space via a non-linear transformation, while the discriminator aims at distinguishing the real examples and the ``fake'' examples synthesized by the generator. The generator gets improved in terms of the capacity of data generation, by learning to deceive the discriminator which also evolves against the generator in such an adversarial learning scheme.

Recently, learning energy-based models (EBMs), with energy functions parameterized by modern convolutional neural networks, for explicit data probability distributions has received significant attention in the fields of  computer vision and machine learning. \citet{XieLuICML,xie2018learning,xie2017synthesizing,xie2019synthesizing,gao2018learning,nijkamp2019learning} suggest that highly realistic examples can be generated by Markov chain Monte Carlo (MCMC) sampling from the learned EBMs. \citet{coopnets2018} propose the Cooperative Networks (CoopNets) framework to learn the EBM simultaneously with a generator model in a cooperative learning scheme, where the generator plays the role of a fast sampler to initialize the MCMC sampling of the EBM, while the EBM teaches the generator via a finite-step MCMC. Within this cooperative learning process, the EBM learns from the training data, while the generator learns from the MCMC sampling of the EBM. In other words, the EBM distills the MCMC into the generator, such that the generator becomes the amortized sampler of the EBM. 
 
Compared to adversarial learning, the energy-based cooperative learning has numerous conceptual advantages for modeling and learning data distribution: (1) \textit{Free of mode collapse}. The training of GANs is known to be difficult, unstable and easy to encounter mode collapse issue \cite{AroraRZ18}. Different from GANs, both EBM and generator in the cooperative learning framework are trained generatively by maximum likelihood. Thus, the CoopNets framework is stable and does not suffer from mode collapse problem \cite{coopnets2018}. (2) \textit{MCMC refinement}. Even though both GAN and CoopNets consist of two sub-models, their interactions in these two frameworks are essentially different. GAN will discard the discriminator once the generator is well-trained, while for the CoopNets, no matter whether at the training stage or testing stage, the EBM enables a refinement for the generator by the iterative MCMC sampling. (3) \textit{Fast-thinking initializer and slow-thinking solver}. Solving a challenging problem usually requires an iterative algorithm. This amounts to slow thinking. However, the iterative algorithm usually needs a good initialization for quick convergence. The initialization amounts to fast thinking. Thus integrating fast-thinking initialization and slow-thinking sampling or optimization is very compelling. \citet{xie2019multimodal} point out that the cooperative learning framework corresponds to a fast-thinking and slow-thinking system, where the generator serves as a fast-thinking initializer and the EBM serves as a slow-thinking solver. The problem we solve in our paper is cross-domain visual translation. 

Our framework for unsupervised cross-domain translation is based on the cooperative learning scheme. We first propose to represent a one-way translator by a cooperative network that includes an energy-based model and a latent variable model, where both of them are trained via MCMC teaching. Specifically, the latent variable model, serving as a translator, maps examples from one domain to another, while the EBM, serving as a teacher, refines the mapped results by MCMC revision, so that the revised results can match to the examples of the target domain in terms of some statistical properties. By simultaneously learning a pair of cooperative networks, each of which is obligated to represent one direction of mapping between two domains, through alternating MCMC teaching, we can achieve a novel framework for unsupervised cross-domain translation. To enforce these two mapping functions to be inverse to each other, we add a cycle consistency loss \cite{zhu2017unpaired} as a constraint to regularize the training of both cooperative networks. This leads to the model we call Cycle-Consistent Cooperative Networks (CycleCoopNets).

Concretely, the contributions of our paper are four-folds: \begin{enumerate}
\item We present a novel energy-based generative framework, CycleCoopNets, to study unsupervised cross-domain translation problem, where we propose to represent a two-way mapping between two domains by a pair of cooperative networks with cycle consistency property, and learn them by the alternating MCMC teaching algorithm. 
\item We apply our framework to a wide range of applications of unsupervised cross-domain translation, including object transfiguration, season transfer, and art style transfer. 
\item We show that our model can achieve competitive quantitative and qualitative results, compared with GAN-based and flow-based \cite{grover2020alignflow} frameworks. 
\item We generalize our framework to the task of unsupervised image sequence translation by combining both spatial and temporal information along with MCMC teaching for appearance translation and motion style preservation. 
\end{enumerate}

\section{Related Work}
Our work is related to the following themes of research.

\textbf{GAN-based cross-domain translation}. Generative Adversarial Networks (GANs) have been successfully applied to a wide range of synthesis problems in the field of computer vision. Three closely related works are Pix2Pix \cite{isola2017image} CycleGAN \cite{zhu2017unpaired} and RecycleGAN \cite{bansal2018recycle}. By generalizing the original unconditioned GANs to the conditioned scenarios, the Pix2Pix is a framework for supervised conditional learning, which has achieved impressive results on paired image-to-image translation tasks, such as image colorization, sketch-to-photo synthesis, etc. CycleGAN is proposed to learn a two-way translator between two domains in the absence of paired examples, by jointly training two GANs, each of which accounts for one-way translation, and enforcing cycle consistency between them. Inspired by CycleGAN, RecycleGAN \cite{bansal2018recycle} is designed to learn a two-way translator between two domains of image sequences without paired examples by adding extra temporal predictive models and enforcing spatiotemporal consistency. Note that our work does not belong to the theme of adversarial learning, even though it also includes encoder-decoder structures like CycleGAN.

\textbf{Energy-based synthesis}. \citet{XieLuICML} propose to adopt a modern convolutional neural network to parameterize the energy function of the energy-based model and learn the model by MCMC-based maximum likelihood estimation. The resulting model is called the generative ConvNet. Compelling results have been achieved by learning the models on images \cite{XieLuICML, du2019implicit, nijkamp2020anatomy}, videos \cite{xie2017synthesizing,xie2019synthesizing}, 3D voxels \cite{xie2018learning, xie20203d} and point clouds \cite{xie2020generative}. \citet{gao2018learning} learn the model with multigid sampling and \citet{nijkamp2019learning} learn the model with short-run MCMC. Our paper is related to energy-based synthesis, because we train energy-based models as teachers for MCMC teaching in our framework.

\textbf{Cooperative learning}. \citet{xie2016cooperative,coopnets2018} proposes the generative cooperative network (CoopNets) that trains an EBM, such as the generative ConvNet model, with the help of a generator network serving as an amortized sampler. The energy-based generative ConvNet distills its knowledge to the generator via MCMC, and this is called MCMC teaching. Recently, \citet{xie2020learning} propose a variant of CoopNets, where a variational auto-encoder (VAE) \cite{kingma2013auto} and an EBM is cooperatively trained via MCMC teaching.  
 \citet{xie2019multimodal} further propose the conditional version of CoopNets model for supervised image-to-image translation. Unlike the above approaches, our framework simultaneously trains two cooperative networks, each of which accounts for one direction of mapping between two domains, and enforces cycle consistency between the two mappings for unsupervised image-to-image translation.

\textbf{Style transfer using neural networks}. \citet{gatys2016image} first propose to use a ConvNet structure, which is pre-trained for image classification, to transfer the artistic style of a style image to a content image. Such a neural style transfer is achieved by synthesizing an image that matches the style of the style image and the content of the content image in terms of Gram matrix statistics of the pre-trained VGG \cite{simonyan2014very} features. Other works include \citet{johnson2016perceptual,ulyanov2016texture,luan2017deep,zhang2018metastyle}. Our paper studies learning a bidirectional mapping between two domains, rather than a unidirectional mapping between two specific instances. Also, our model can be applied to not only style transfer but also other image-to-image translation tasks, e.g., object transfiguration, etc.

\section{Proposed Framework}
\label{cycle_coopnet}
\subsection{Problem Definition}
Suppose we have two different domains, say $\x$ and $\y$, and two data collections from these domains $\{x_i, i=1,...,n_x\}$ and $\{y_i, i=1,...,n_y\}$, where $x_i \in \x$ and $y_i \in \y$. $n_x$ and $n_y$ are numbers of examples in the collections, respectively. $n_x$ and $n_y$ are not necessarily the same. Let $p_{\text{data}}(x)$ and $p_{\text{data}}(y)$ be the unknown data distributions of these two domains. Without instance-level correspondence between two collections, we want to learn mapping functions between two domains for the sake of cross-domain translation. 

\subsection{Latent Variable Model as a Translator} 

Let us talk about one-way translation problem first. To transfer image across domains, say $\y$ to $\x$, we specify a  mapping $G_{\y \rightarrow \x}$ that seeks to re-express the image $y \in \y$ by the image $x \in \x$. The latent variable model is of the form:  
\begin{eqnarray}
\begin{aligned} 
&y \sim p_{\rm data}(y),  \\  
&x = G_{\y \rightarrow \x}(y; \alpha_{\x}) + \epsilon,  \epsilon \sim \mathcal{N}(0, \sigma^2  I_D), \label{eq:mapping}
\end{aligned} 
\end{eqnarray}
where $G_{\y \rightarrow \x}(y; \alpha_{\x})$ is parameterized by an encoder-decoder structure whose parameters are denoted by $\alpha_{\x}$, and $\epsilon$ is a Gaussian residual. We assume $\sigma$ is given and $I_D$ is the $D$-dimensional identity matrix. In model (\ref{eq:mapping}), $y$ is the latent variable of $x$, because for each $x \in \x$, its version $y \in \y$ is unobserved. ($x$ and $y$ have the same number of dimensions.)

Given the empirical prior distribution $p_{\rm data}(y)$ and $q(x|y;\alpha_{\x}) \sim \mathcal{N}(G_{\y \rightarrow \x}(y; \alpha_{\x}), \sigma^2  I_D)$, we can get the joint density $q(x,y; \alpha_{\x})=p_{\rm data}(y)q(x|y;\alpha_{\x})$, and the marginal density $q(x; \alpha_{\x})= \int q(x,y; \alpha_{\x}) dy$. Training the model via maximum likelihood estimation (MLE) requires the prior $p_{\rm data}(y)$ to be a tractable density (e.g., Gaussian white noise distribution) for calculating the derivative of the data log-likelihood with respect to $\alpha_{\x}$, i.e., $\frac{\partial}{\partial \alpha_{\x}} [\frac{1}{n_x} \sum_{i=1}^{n_x} \log q(x_i;\alpha_{\x})]$, from training examples $\{x_i,i=1,...,n_{\x}\}$ in domain $\x$. Due to the unknown prior $p_{\rm data}(y)$, we can not estimate $\alpha_{\x}$ in model (\ref{eq:mapping}) via MLE with an explaining away inference \cite{han2017alternating} or a variational inference \cite{kingma2013auto}.

\subsection{Energy-Based Model as a Teacher}
 Instead, to avoid the challenging problem of inferring $y$ from $x$ in estimating $\alpha_{\x}$, we can train $G_{\y \rightarrow \x}$ via MCMC teaching by recruiting an energy-based model (EBM), such as the generative ConvNet \cite{XieLuICML}, which specifies the distribution of $x$ explicitly up to a normalizing constant:
\begin{eqnarray} 
p(x; \theta_{\x}) = \frac{1}{Z(\theta_{\x})} \exp\left[ f(x; \theta_{\x})\right]p_{0}(x) \label{model:uncondi},
\end{eqnarray}  
where $Z(\theta_{\x})=\int \exp \left[f(x;\theta_{\x}) \right]p_{0}(x)dx$ is the intractable normalizing constant, $p_{0}(x)$ is the Gaussian reference distribution, i.e., $p_{0}(x)\propto\exp(-||x||^2/2s^2)$. Standard deviation $s$ is a hyperparameter. The energy function $\mathcal{E}(x;\theta_{\x})=-f(x;\theta_{\x})+ ||x||^2/2s^2$, where $f$ is parameterized by a ConvNet with parameters $\theta_{\x}$.  As to learning $p(x; \theta_{\x})$, the maximum likelihood estimator equivalently minimizes the Kullback-Leibler (KL) divergence between the data distribution $p_{\text{data}}(x)$ and the model, $\KL(p_{\text{data}}(x)\|p(x;\theta_{\x}))$, over $\theta_{\x}$. The gradient of MLE is given by
\begin{eqnarray} 
\begin{aligned}
&- \frac{\partial}{\partial \theta_{\x}} {\rm KL}(p_{\rm data}(x)\| p(x;\theta_{\x})) \label{eq:eb0} \\
= &\E_{p_{\rm data}} \left[\frac{\partial}{\partial \theta_{\x}} f(x;\theta_{\x})\right] -  \E_{p_{\theta_{\x}}} \left[\frac{\partial}{\partial \theta_{\x}} f(x;\theta_{\x}) \right], \label{eq:eb}
\end{aligned}
\end{eqnarray} 
where $\E_{p_{\theta_{\x}}}$ denotes the expectation with respect to the model $p(x; \theta_{\x})$. 
In practice, Eq.(\ref{eq:eb}) can be approximated by 
\begin{eqnarray} 
\Delta(\theta_{\x}) = \frac{1}{n} \sum_{i=1}^{n} \frac{\partial}{\partial \theta_{\x}} f(x_i; \theta_{\x}) - \frac{1}{n} \sum_{i=1}^{\tilde{n}} \frac{\partial}{\partial \theta_{\x}} f(\tilde{x}_i; \theta_{\x})  \label{eq:learningD}
\end{eqnarray}
where $\{\tilde{x}_i, i=1,...,\tilde{n}\}$ are MCMC examples sampled from the current distribution $p(x; \theta_{\x})$. 
With an EBM model defined on domain $\x$, we can sample $x$ by MCMC, such as Langevin dynamics \cite{neal2011mcmc}, which iterates
\begin{eqnarray}
   x_{\tau+1} = x_\tau - \frac{\delta^2}{2}  \frac{\partial}{\partial x} \mathcal{E}(x_\tau; \theta_{\x})  + \delta U_{\tau},  \label{eq:LangevinD}
\end{eqnarray}  
where $\tau$ indexes the time step, $\delta$ is the step size, and $U_{\tau} \sim \mathcal{N}(0,I_D)$ is the Gaussian noise term. 

In the MCMC teaching process, the EBM $p(x; \theta_{\x})$ can distill its MCMC algorithm to $q(x;\alpha_{\x})$, so that $p(x;{\theta_{\x}})$ can absorb and accumulate the MCMC transitions in order to reproduce them by one step ancestral sampling. That means, at each MCMC teaching step, $q(x;\alpha_{\x})$ learns to get close to $p(x; \theta_{\x})$, and chases it toward $p_{\rm data}(x)$. Meanwhile, $q(x;\alpha_{\x})$ will serve as an initializer of the MCMC of $p(x; \theta_{\x})$ for efficient Langevin sampling. 

Formally, let $\M_{\theta_{\x}}$ be the Markov transition kernel of $l$ steps of Langevin dynamics that samples from $p(x;\theta_{\x})$, and $\M_{\theta_{\x}} q_{\alpha_{\x}}$ be the marginal distribution obtained by running the Markov transition $\M_{\theta_{\x}}$ initialized by $q_{\alpha_{\x}}$. Training $q_{\alpha_{\x}}$ via MCMC teaching seeks to find $\alpha_{\x}$ at time $t$ to minimize $\KL(\M_{\theta_{\x}} q_{\alpha_{\x}^{(t)}}||q_{\alpha})$, which implies a minimization of $\KL(p_{\theta_{\x}}||q_{\alpha_{\x}})$ over $\alpha_{\x}$. Once $\KL(p_{\rm data}(x)||p(x;\theta_{\x}))\rightarrow 0$, then  $\KL(p_{\rm data}(x)||q(x;\alpha_{\x}))\rightarrow 0$.

\subsection{Cycle-Consistent Cooperative Networks}
The energy-based model $p$ and the latent variable model $G$ form a cooperative network. In this paper, to tackle the unsupervised cross-domain translation, we propose a framework consisting of a pair of cooperative networks, i.e., 
\begin{eqnarray}
\begin{aligned}
\y \rightarrow \x &: \{G_{\y \rightarrow \x}(y;\alpha_{\x}), p(x;\theta_{\x})\},\\
\x \rightarrow \y &: \{G_{\x \rightarrow \y}(x;\alpha_{\y}), p(y;\theta_{\y})\},\\
\end{aligned}
\end{eqnarray}
each of which accounts for one direction of translation between domains $\x$ and $\y$, and we simultaneously learn them by alternating their MCMC teaching algorithms. To guarantee that each individual input $x \in \x$ or $y \in \y$ and its translated version $\tilde{y} \in \y$ or $\tilde{x} \in \x$ are meaningfully paired up, we enforce $G_{\y \rightarrow \x}$ and $G_{\x \rightarrow \y}$ to be inverse functions of each other when training the models, i.e.,
\begin{eqnarray}
\begin{aligned}
x = G_{\y \rightarrow \x}(G_{\x \rightarrow \y}(x; \alpha_{\y});\alpha_{\x}) , \forall x,\\
y = G_{\x \rightarrow \y}(G_{\y \rightarrow \x}(y; \alpha_{\x});\alpha_{\y}) , \forall y.\\
\end{aligned}
\end{eqnarray}
We call the proposed framework Cycle-Consistent Cooperative Networks (CycleCoopNets).

\subsection{Alternating MCMC Teaching}

\begin{figure}[t]
\centering
\includegraphics[width=0.95\linewidth]{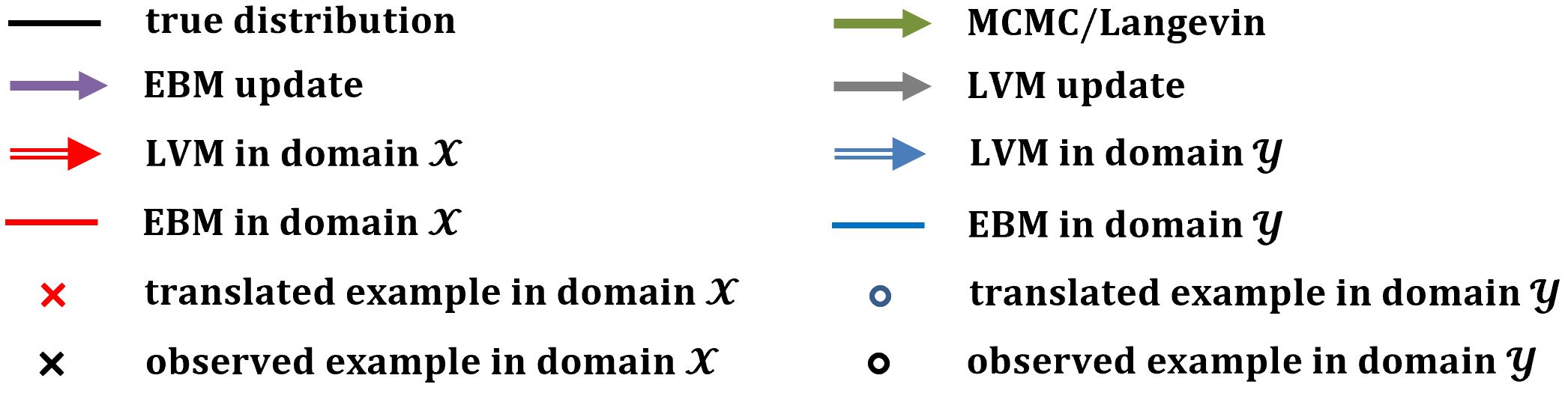}
\includegraphics[width=0.95\linewidth]{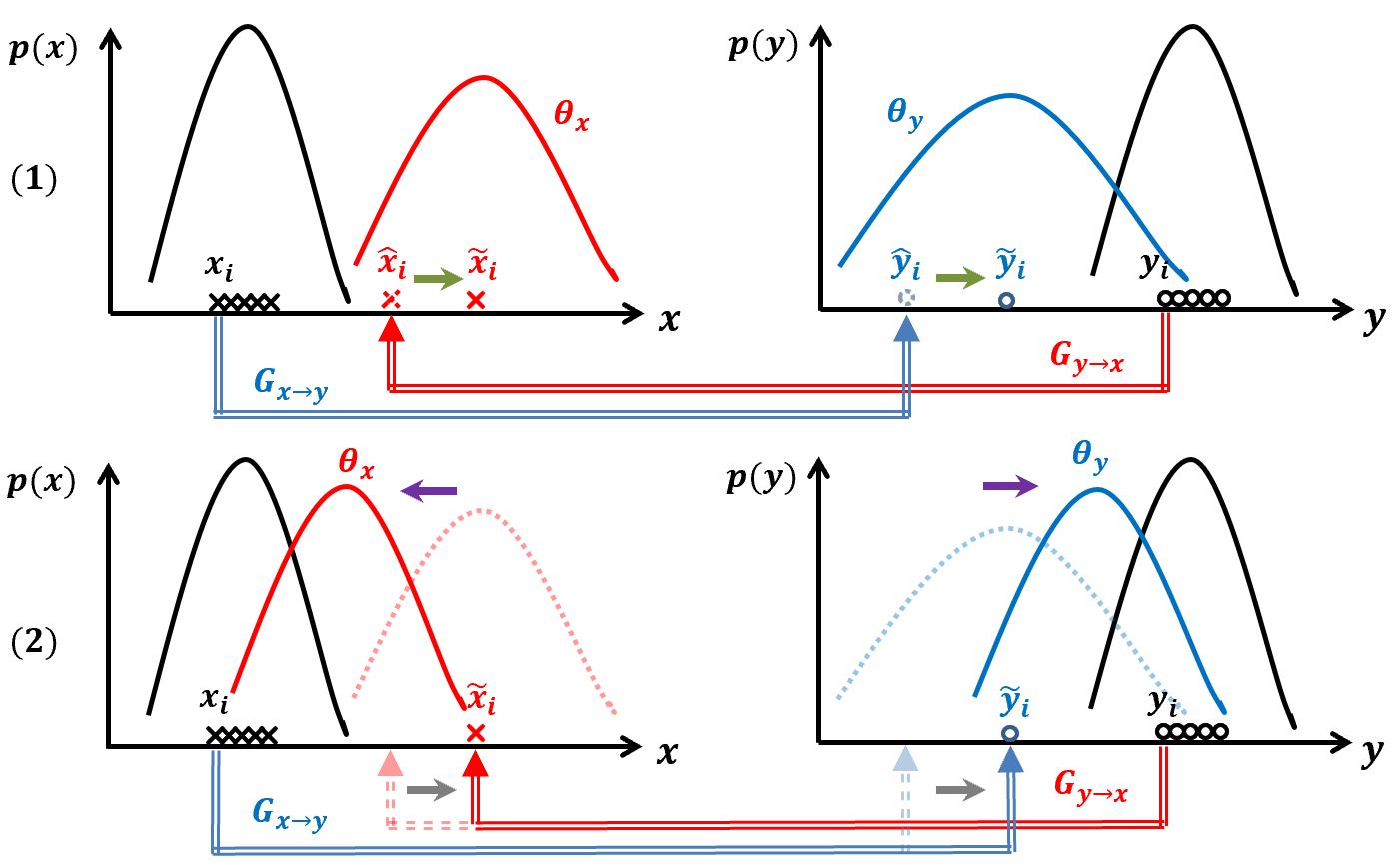}
\includegraphics[width=0.95\linewidth]{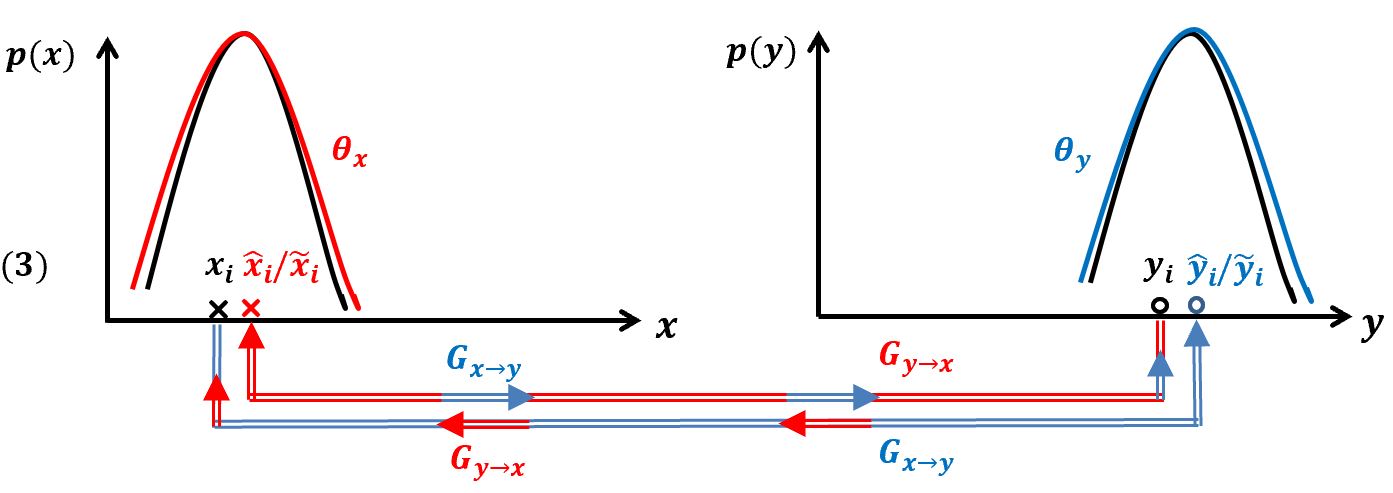}
\caption{An illustration of the alternating MCMC teaching algorithm. (1) cross-domain mapping (2) density shifting (3) mapping shifting with cycle consistency.}\label{fig:alter_mcmc_teaching}
\end{figure}

As illustrated in Figure \ref{fig:alter_mcmc_teaching}(1), we first sample $y_i \sim p_{\text{data}}(y)$, and then translate via $\hat{x}_i=G_{\y \rightarrow \x}(y_i;\alpha_{\x})$, for $i=1,...,\tilde{n}$. Starting from $\{\hat{x}_i, i=1,...,\tilde{n}\}$, we run MCMC, e.g., Langevin dynamics, for a finite number of steps toward $p(x;\theta_{\x})$ to obtain $\{\tilde{x}_i, i=1,...,\tilde{n}\}$, which are revised versions of $\{\hat{x}_i, i=1,...,\tilde{n}\}$. Even though $\{\tilde{x}_i\}$ are cooperatively generated by both $p(x;\theta_{\x})$ and $G_{\y \rightarrow \x}$, they are synthesized examples sampled from $p(x;\theta_{\x})$, and are used to learn  $\theta_{\x}$ according to Eq.(\ref{eq:learningD}). After updating $\theta_{\x}$, $p(x;\theta_{\x})$ gets close to $p_{\rm data}(x)$ by fitting all the major modes of $p_{\rm data}(x)$. See Figure \ref{fig:alter_mcmc_teaching}(2) for an explanation. The energy-based model $p(x;\theta_{\x})$ then teaches $G_{\y \rightarrow \x}$ by MCMC teaching. The key is that for the cooperatively synthesized examples $\{\tilde{x}_i\}$, their sources  $\{y_i\}$ are known. In order to update $\alpha_{\x}$ of $G_{\y \rightarrow \x}$, we treat $\{\tilde{x}_i\}$ as the training data of $G_{\y \rightarrow \x}$. Since these $\{\tilde{x}_i\}$ are obtained by the Langevin dynamics initialized from $\{\hat{x}_i\}$, which are generated by $G_{\y \rightarrow \x}$ with known inputs $\{y_i\}$, we can directly update $\alpha_{\x}$ by learning from the complete data $\{(y_i, \tilde{x}_i), i = 1,...,\tilde{n}\}$, which is a non-linear regression of $\tilde{x}_i$ on $y_i$ with the objective
\begin{eqnarray}
L_{teach}(\alpha_{\x})= \frac{1}{\tilde{n}}\sum_{i=1}^{\tilde{n}}\|\tilde{x}_i - G_{\y \rightarrow \x}(y_i;\alpha_{\x})\|^2.  \label{eq:G_loss}
\end{eqnarray}
At $\alpha_{\x}^{(t)}$, $y_i$ is mapped to the initial example $\hat{x}_i$. After updating $\alpha_{\x}$, we want $y_i$ to map the revised example $\tilde{x}_i$. That is, we revise $\alpha_{\x}$ to absorb the MCMC transition from $\hat{x}_i$ to $\tilde{x}_i$ for chasing 
$p(x;\theta_{\x})$. In the meanwhile, we simultaneously train the other mapping from domain $\x$ to $\y$, i.e., $\{G_{\x \rightarrow \y}, p(y;\theta_{\y})\}$ in a similar way. To enforce mutual invertibility between $G_{\x \rightarrow \y}$ and $G_{\y \rightarrow \x}$, we can add the following cycle consistency loss while learning $\alpha_{\x}$ and $\alpha_{\y}$, 
\begin{align} 
 L_{\rm cyc}&(\alpha_{\x},\alpha_{\y})=\frac{1}{\tilde{n}} \sum_{i=1}^{\tilde{n}}\|x_i - G_{\y \rightarrow \x}(G_{\x \rightarrow \y}(x_i; \alpha_{\y});\alpha_{\x})\|_1 \nonumber \\ 
 &+ \frac{1}{\tilde{n}}\sum_{i=1}^{\tilde{n}}\|y_i - G_{\x \rightarrow \y}(G_{\y \rightarrow \x}(y_i; \alpha_{\x});\alpha_{\y})\|_1.
 \label{eq:inv_loss}
\end{align} 
Figure \ref{fig:alter_mcmc_teaching}(3) is an illustration of the convergence of the algorithm. Algorithm \ref{code:3} presents a full description of the learning algorithm to train CycleCoopNets. 

\section{Generalizing to Unpaired Cross-Domain Image Sequence Translation}

We can further generalize the proposed framework to learning a translation between two domains of image sequences where paired examples are unavailable. For example, given an image sequence of Donald Trump's speech, we can translate it to an image sequence of Barack Obama, where the content of Donald Trump is transferred to Barack Obama but the speech is in Donald Trump's style. Such an appearance translation and motion style preservation framework may have a wide range of applications in video manipulation. Even though the translation is made on image sequences, we do not build distributions or mappings on sequence domain. Instead, we rely on translation models defined on image space, and bring in temporal prediction models accounting for temporal information. Therefore, we can make minimal modifications on our current framework discussed in Section \ref{cycle_coopnet} for image sequence translation. Suppose we observe two unpaired but ordered image sequences $X=(x_1, x_2,...,x_t,...)$  and $Y=(y_1, y_2,...,y_t,...)$, and $\forall x_t \in \x, \forall y_t \in \y$. Each long sequence can be turned into a collection of short sequences with an equal length, i.e., $\{x_{t:t+k}\}_{t=1}^{T_\x}$ and $\{y_{t:t+k}\}_{t=1}^{T_\y}$, where $x_{t:t+k}=(x_t,...,x_{t+k})$ and $y_{t:t+k}=(y_t,...,y_{t+k})$. The length of each short sequence is $k+1$. The current framework only learns translation of static image frames between two domains, without considering temporal information existing in each domain. We need to make the following two modifications to adapt our model to the new task: (i) We learn a temporal prediction model in each domain to predict future image frame given the past image frames in a sequence. Let $R_{\x}$ and $R_{\y}$ denote temporal prediction models for domains $\x$ and $\y$ respectively.  We learn $R_{\x}$ and $R_{\y}$ by minimizing
\begin{eqnarray}
\begin{aligned}
   L_{\rm tp}(R_{\x}) &=\frac{1}{T_{\x}} \sum_{t=1}^{T_{\x}} \| x_{t+k} - R_{\x} (x_{t:t+k-1}) \|_1,\\
   L_{\rm tp}(R_{\y}) &=\frac{1}{T_{\y}} \sum_{t=1}^{T_{\y}} \| y_{t+k} - R_{\y} (y_{t:t+k-1}) \|_1. 
   \label{eq:recurrent_loss}
   \end{aligned}
\end{eqnarray} 
(ii) With $R_{\x}$ and $R_{\y}$, we can modify the loss in Eq.~(\ref{eq:inv_loss}) to take into account spatial-temporal information as below
\begin{eqnarray} 
\begin{aligned}
 & L_{\rm st}(G_{\x \rightarrow \y}, R_{\y}, G_{\y \rightarrow \x}) \\
 = & \frac{1}{T_{\x}} \sum_{t=1}^{T_\x} \|x_{t+k} - G_{\y \rightarrow \x}(R_{\y}(G_{\x \rightarrow \y}(x_{t:t+k-1})))\|_1, \\
  & L_{\rm st}(G_{\y \rightarrow \x}, R_{\x}, G_{\x \rightarrow \y}) \\
  =& \frac{1}{T_{\y}} \sum_{t=1}^{T_{\y}} \|y_{t+k} - G_{\x \rightarrow \y}(R_{\x}(G_{\y \rightarrow \x}(y_{t:t+k-1})))\|_1,  
\label{eq:recycle_loss}
\end{aligned}
\end{eqnarray} 
where, for notation simplicity, we use $G_{\x \rightarrow \y}(x_{t:t+k-1})=(G_{\x \rightarrow \y}(x_t),...,G_{\x \rightarrow \y}(x_{t+k-1}))$ and  $G_{\y \rightarrow \x}(y_{t:t+k-1})=(G_{\y \rightarrow \x}(y_t),...,G_{\y \rightarrow \x}(y_{t+k-1}))$. The final objective of $G$ and $R$ is given by
\begin{eqnarray}
\begin{aligned}
\min_{G,R}& L(G,R) =  L_{\rm teach}(G_{\y \rightarrow \x}) + L_{\rm teach}(G_{\x \rightarrow \y})\\
& +\lambda_1 L_{\rm tp}(R_{\x}) +\lambda_2 L_{\rm st}(G_{\x \rightarrow \y}, R_{\y}, G_{\y \rightarrow \x}) \\
&+ \lambda_1 L_{\rm tp}(R_{\y}) + \lambda_2 L_{\rm st}(G_{\y \rightarrow \x}, R_{\x}, G_{\x \rightarrow \y}),  \label{eq:final_loss}
\end{aligned}
\end{eqnarray}
where $\lambda_1$ and $\lambda_2$ are hyper-parameters. During testing, given a testing image sequence from domain $\x$, $(x_1, x_2,...,x_t,...)$, we can translate the whole sequence to domain $\y$ by mapping each image frame $x_t$ to domain $\y$ via the learned $G_{\x \rightarrow \y}$, and then revising the result via $p(y;\theta_{\y})$.

\begin{algorithm}[t]
\caption{Alternating MCMC teaching algorithm}
\label{code:3}
\begin{algorithmic}[1]

\REQUIRE
\STATE (1) training examples in domain $\x$, $\{x_i, i=1,...,n_x\}$, and domain $\y$, $\{y_i, i=1,...,n_y\}$; (2) number of Langevin steps $l$; (3) learning rate $\gamma_{\theta_{\x}}, \gamma_{\theta_{\y}}, \gamma_{\alpha_{\x}}, \gamma_{\alpha_{\y}}$; (4) number of learning iterations $T$.

\ENSURE
\STATE Estimated parameters $\theta_{\x},\theta_{\y}$, $\alpha_{\x}$, $\alpha_{\y}$

\item[]
\STATE Let $t\leftarrow 0$, randomly initialize $\theta_{\x},\theta_{\y}$, $\alpha_{\x}$, $\alpha_{\y}$
\REPEAT 
\STATE  $\{y_i \sim p_{data}(y)\}_{i=1}^{\tn}$
\STATE  $\{\hat{x}_i = G_{\y \rightarrow \x}(y_i; \alpha_{\x})\}_{i=1}^{\tn}$
\STATE $\{x_i \sim p_{data}(x)\}_{i=1}^{\tn}$ \STATE $\{\hat{y}_i = G_{\x \rightarrow \y}(x_i; \alpha_{\y})\}_{i=1}^{\tn}$.
\STATE Starting from $\{\hat{x}_i\}_{i=1}^{\tn}$, run $l$ steps of Langevin revision in Eq.~(\ref{eq:LangevinD}) to obtain $\{\tilde{x}_i\}_{i=1}^{\tn}$.
\STATE Starting from $\{\hat{y}_i\}_{i=1}^{\tn}$, run $l$ steps of Langevin revision in Eq.~(\ref{eq:LangevinD}) to obtain $\{\tilde{y}_i\}_{i=1}^{\tn}$. 
\STATE Given $\{x_i\}_{i=1}^{\tn}$ and $\{\tilde{x}_i\}_{i=1}^{\tn}$, update $\theta_{\x}^{(t+1)} = \theta_{\x}^{(t)} +  \gamma_{\theta_{\x}}\Delta(\theta_{\x}^{(t)})$,  where $\Delta(\theta_{\x}^{(t)})$ is computed by Eq.~(\ref{eq:learningD}). 
\STATE Given $\{y_i\}_{i=1}^{\tn}$ and $\{\tilde{y}_i\}_{i=1}^{\tn}$, update $\theta_{\y}^{(t+1)} = \theta_{\y}^{(t)} + \gamma_{\theta_{\y}} \Delta(\theta_{\y}^{(t)})$,  where $\Delta(\theta_{\y}^{(t)})$ is computed by Eq.~(\ref{eq:learningD}). 
\STATE Given $\{y_i\}_{i=1}^{\tn}$, $\{x_i\}_{i=1}^{\tn}$ and $\{\tilde{x}_i\}_{i=1}^{\tn}$, update $\alpha_{\x}^{(t+1)} = \alpha_{\x}^{(t)} -  \gamma_{\alpha_{\x}} \nabla (\alpha_{\x}^{(t)})$,  where $\nabla(\alpha_{\x}^{(t)})$ is the gradient of loss functions in Eq.~(\ref{eq:G_loss}) and Eq.~(\ref{eq:inv_loss}). 
\STATE Given $\{x_i\}_{i=1}^{\tn}$, $\{y_i\}_{i=1}^{\tn}$ and $\{\tilde{y}_i\}_{i=1}^{\tn}$, update $\alpha_{\y}^{(t+1)} = \alpha_{\y}^{(t)} -  \gamma_{\alpha_{\y}} \nabla(\alpha_{\y}^{(t)})$,  where $\nabla(\alpha_{\y}^{(t)})$ is the gradient of loss functions in Eq.~(\ref{eq:G_loss}) and Eq.~(\ref{eq:inv_loss}).   
\STATE Let $t \leftarrow t+1$
\UNTIL $t = T$
\end{algorithmic}
\end{algorithm}

\section{Experiments}

We perform experiments on the tasks of unsupervised image-to-image translation and image sequence translation to evaluate the CycleCoopNets. 
The code can be found at the page
\url{http://www.stat.ucla.edu/~jxie/CycleCoopNets/}.

\subsection{Unsupervised Image-to-Image Translation}	\label{sec:image}

\subsubsection{Implementation} We present the network structures of $p$ and $G$ for the mapping from domain ${\cal Y}$ to ${\cal X}$ and the mapping from domain ${\cal X}$ to ${\cal Y}$ as below. We use the same bottom-up network structures to parameterize the negative energy functions $f$ for both $p(x;\theta_{\x})$ and $p(y;\theta_{\y})$, and also the same encoder-decoder structures for $G_{\x\rightarrow\y}$ and $G_{\y\rightarrow\x}$.

\textit{Structure of $p$}: Each EBM $p$ has a bottom-up ConvNet structure $f$ that consists of 4 layers of convolutions with numbers of channels $\{64, 128, 256, 512\}$, filter sizes $\{3, 4, 4, 4\}$, and subsampling factors $\{1, 2, 2, 2\}$ at different layers, and one fully connected layer with 100 filters. Leaky ReLU layers are applied between convolutional layers.

\textit{Structure of $G$}. We adopt the architecture from \citet{johnson2016perceptual} for $G$. The CycleGAN also uses the same architecture. We use 9 residual blocks. We only replace the instance normalization by the batch normalization in this architecture for art style transfer. 

\textit{Hyperparameters} We use 20 Langevin steps for season transfer and 15 steps for other experiments. The Langevin step size is 0.002. The standard deviation $s$ of the reference distribution of the EBM is 0.016. We use Adam \cite{kingma2015adam} for optimization with a learning rate 0.0002. The hyperparameter that weighs the relative contribution of the cycle consistency loss term in the total loss is $\lambda_{cyc}=9$. The batch size is 1. The number of parallel chains is 1.

\begin{figure}[t]
\setlength{\tabcolsep}{1pt}
\centering
\centering
\begin{tabular}{ccccc}
Input & CycleGAN & UNIT & \textbf{$G$} (ours) & \textbf{$p$} (ours) \\
\includegraphics[width=.19\linewidth]{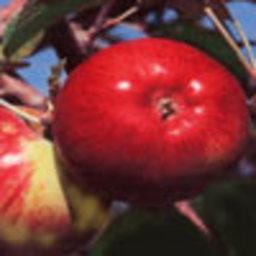} &
\includegraphics[width=.19\linewidth]{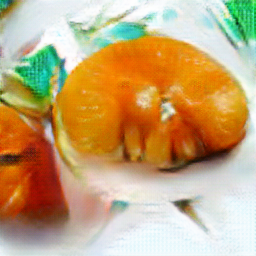} &
\includegraphics[width=.19\linewidth]{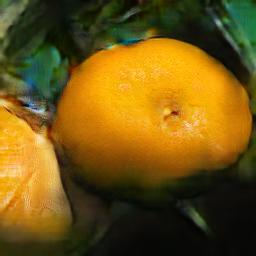} &
\includegraphics[width=.19\linewidth]{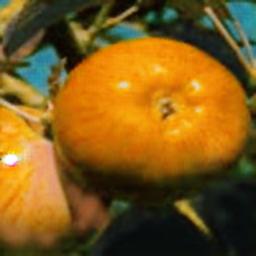} &
\includegraphics[width=.19\linewidth]{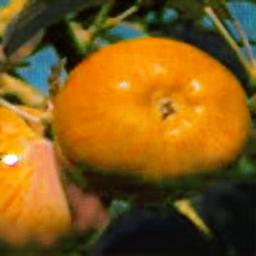} \\
\includegraphics[width=.19\linewidth]{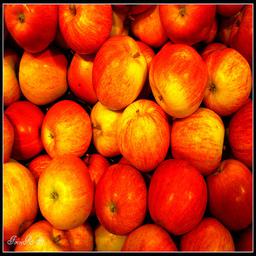} &
\includegraphics[width=.19\linewidth]{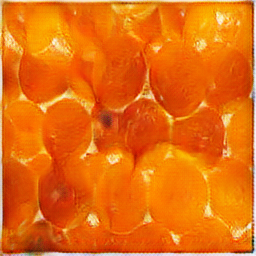} &
\includegraphics[width=.19\linewidth]{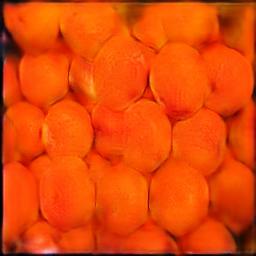} &
\includegraphics[width=.19\linewidth]{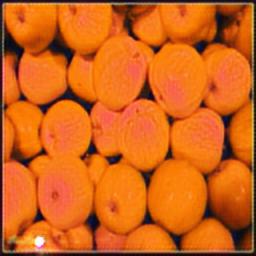} &
\includegraphics[width=.19\linewidth]{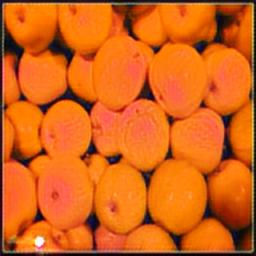} 
\end{tabular}
apple $\Rightarrow$ orange
\begin{tabular}{ccccc}
\\
Input & CycleGAN & UNIT & \textbf{$G$} (ours) & \textbf{$p$} (ours) \\
\includegraphics[width=.19\linewidth]{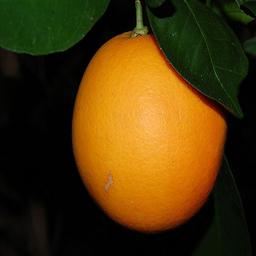} &
\includegraphics[width=.19\linewidth]{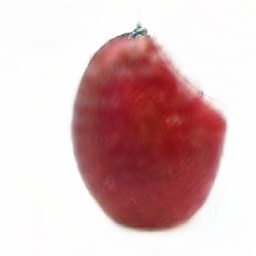} &
\includegraphics[width=.19\linewidth]{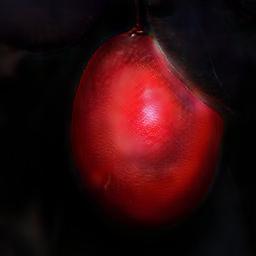} &
\includegraphics[width=.19\linewidth]{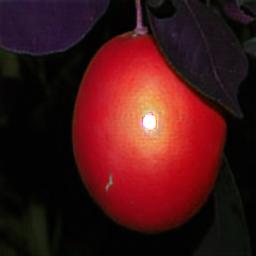} &
\includegraphics[width=.19\linewidth]{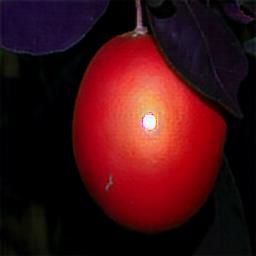} \\
\includegraphics[width=.19\linewidth]{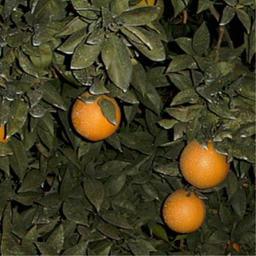} &
\includegraphics[width=.19\linewidth]{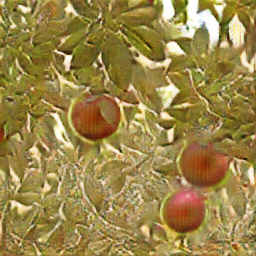} &
\includegraphics[width=.19\linewidth]{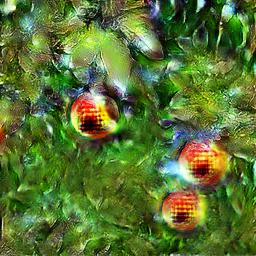} &
\includegraphics[width=.19\linewidth]{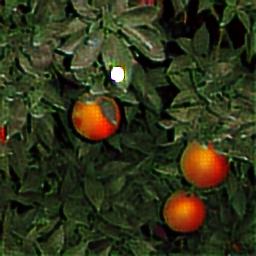} &
\includegraphics[width=.19\linewidth]{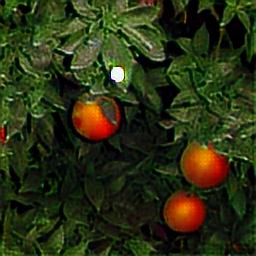}
\end{tabular}
orange $\Rightarrow$ apple
\caption{Object transfiguration. The top panel displays the translation from apples to oranges, and the bottom panel displays the translation from oranges to apples. For each panel, the first column shows the input images, and the rest show the translated results obtained by different models.}
\label{apple2orange}
\end{figure}

\subsubsection{Object transfiguration} We train our model to translate one object category from ImageNet \cite{deng2009imagenet} to another. Each category has roughly 1,000 training examples. Figure \ref{apple2orange} displays some testing results of an example of object transfiguration between categories apple and orange. Each panel shows one direction of translation, in which the first column displays the input images, the second and the third columns show the results obtained by two baseline methods CycleGAN \cite{zhu2017unpaired} and UNIT \cite{liu2017unsupervised}, respectively. The last two columns show the results achieved by our model, where the fourth column displays the results generated by $G$ without using $p$'s MCMC revision, and the fifth column displays those obtained by $p$'s MCMC, which is initialized by $G$. These qualitative results suggest that the proposed framework can be successfully applied to unsupervised image-to-image translation. Moreover, for each pair of $G$ and $p$, $G$ traces $p$, and $p$ traces $p_{\text{data}}$, thus $G$ will eventually get close to $p$. The results shown in Figure \ref{apple2orange} verify this fact in the sense that the final results generated by $G$ (fourth column) and $p$ (fifth column) look almost the same and indistinguishable.    

\begin{table}[ht!]
\centering
\begin{center}

\begin{tabular}[u]{c|cccc}

\hline
\multirow{2}{*}{methods} & \multicolumn{2}{c}{\textbf{apple $\Rightarrow$ orange}} & \multicolumn{2}{c}{\textbf{orange $\Rightarrow$ apple}} \\
 & FID $\downarrow$ & DIPD $\downarrow$ & FID$\downarrow$ & DIPD $\downarrow$ \\ \hline
CycleGAN &  160.78 & 1.75 & 143.87 & 1.73 \\ 
UNIT & 170.66  & 1.58 & 122.04 & 1.62 \\
$G (l=15)$ &   158.66 & 1.28 & 119.27 & 1.34 \\ 
$p (l=15)$ &  \textbf{154.58}   & \textbf{1.23} & \textbf{118.82} & \textbf{1.25} \\ \hline
$p (l=1)$ &   192.60 & 1.43 & 143.00 & 1.42 \\ 
$p (l=5)$&   166.41 & 1.43 & 170.38 & 1.40 \\ 
$p (l=10)$ &   189.60 & 1.32 & 141.60 & 1.32 \\ \hline
\end{tabular}
 \caption{ Quantitative evaluation on apple$\Leftrightarrow$orange dataset with respect to Fr\'echet Inception Distance (FID) and Domain-invariant Perceptual Distance (DIPD). The top two rows show the results of CycleGAN and UNIT. The middle two rows show the results of $G$ and $p$, where $l$ is the number of Langevin steps. The last three rows show performances of the models with different numbers of Langevin steps.}
\label{tab:FID}
\end{center}
\end{table}

\begin{figure}[ht!]
\centering
\includegraphics[width=.71\linewidth]{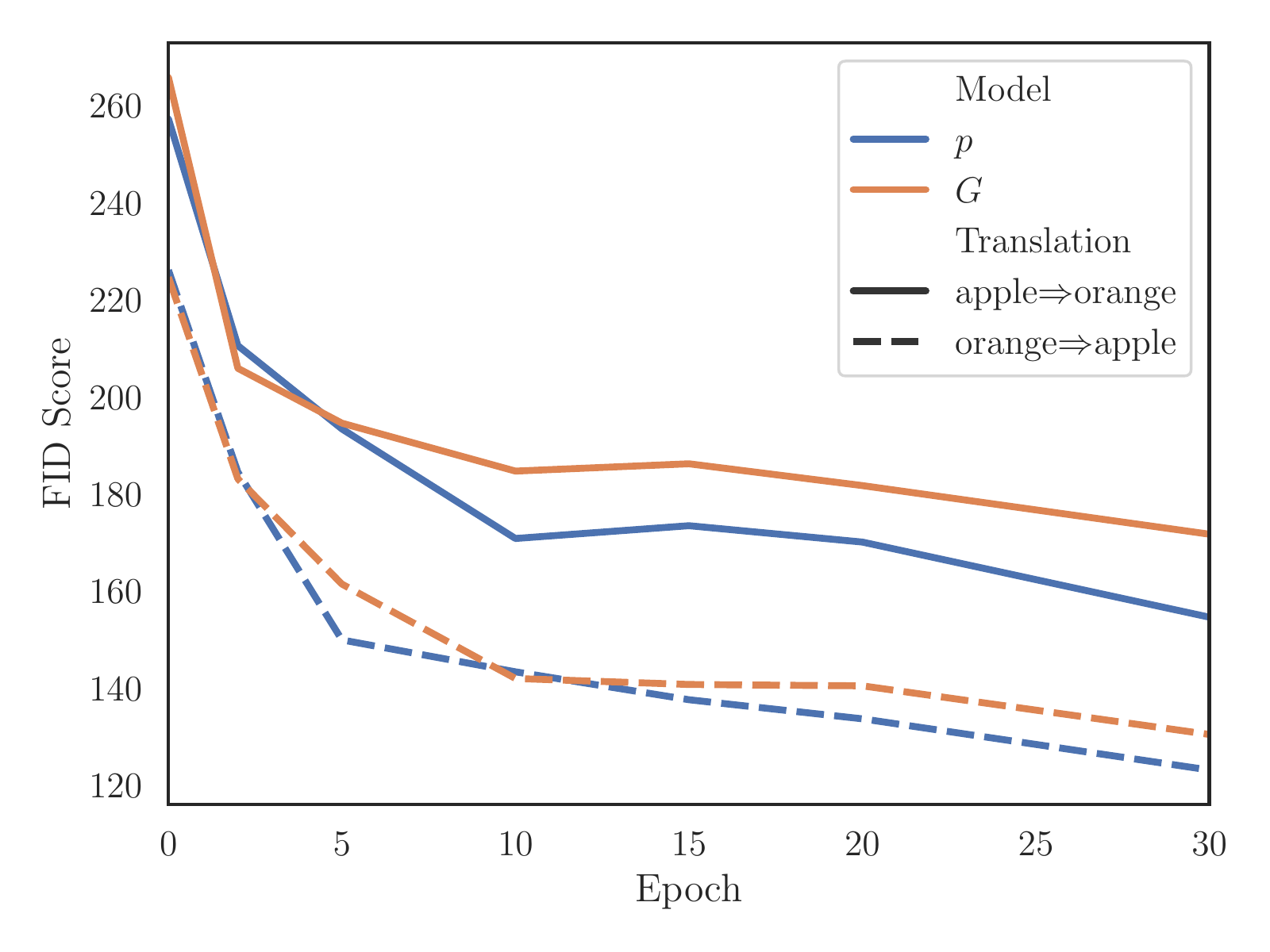}
\caption{Learning curves that represent FID scores of the translated images (apple $\Rightarrow$ orange or orange $\Rightarrow$ apple) over epochs. 
The blue and yellow curves represent the result achieved by $p$ and $G$ respectively. 
The results suggest that $p$ refines the results provided by $G$.  }
\label{fig:curve_fid_FID}
\end{figure}

To quantitatively evaluate the performance of the proposed model, we use the Fr\'echet Inception Distance
(FID) \cite{heusel2017gans} to measure the similarity between the translated distribution and the target distribution. This distribution matching metric can indicate to what extent the input images in one domain are translated to the other domain. We compute the FID score between the set of translated images and the set of observed images in the target domain. We use the activations from the last average pooling layer of the Inception-V3 \cite{szegedy2016rethinking} model, which is pretrained on ImageNet \cite{deng2009imagenet} for classification, as features of each image for computing the FID. A lower FID score is desired because it corresponds to a higher similarity between the target distribution and the translated one.  

We additionally evaluate our results by the domain-invariant perceptual distance (DIPD)~\cite{huang2018multimodal}, which can be used to measure the content preservation in unsupervised image-to-image translation. According to \citet{huang2018multimodal}, the DIPD is given by the L2-distance between the normalized VGG \cite{simonyan2014very} Conv5 features of the input image and the translated image. We expect the content in the input image is preserved in the translated image, thus a lower DIPD is desired.  

As shown in Table \ref{tab:FID}, the proposed framework outperforms the baseline models CycleGAN and UNIT on both metrics FID and DIPD. 
We also study the effect of the number of Langevin steps used in the model. Performances of our models with different numbers of Langevin steps are evaluated by FID and DIPD in the last 3 rows of Table \ref{tab:FID}.

Figure \ref{fig:curve_fid_FID} shows the learning curves that represent FID scores of the translated images obtained by $G$ (orange curves) and $p$ (blue curves) over training epochs. The solid curves represent the results obtained on the translation from apples to oranges, while the dashed curves represent the results for the other direction of translation. We observe improvements in quality of the results in terms of FID score, as the learning algorithm proceeds. We also observe the MCMC refinement effect of $p$ on $G$ in the learning curves.

\begin{figure}[t!]
\captionsetup[subfigure]{labelformat=empty}
\setlength{\tabcolsep}{1pt}

\centering
\begin{tabular}{ccccc}
Input & CycleGAN & UNIT & DRIT & ours \\
\includegraphics[width=.19\linewidth]{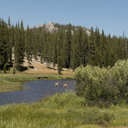} &
\includegraphics[width=.19\linewidth]{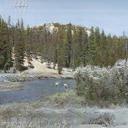} &
\includegraphics[width=.19\linewidth]{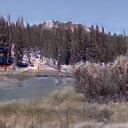} &
\includegraphics[width=.19\linewidth]{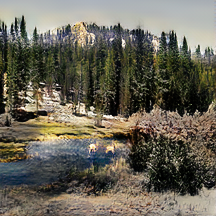} &
\includegraphics[width=.19\linewidth]{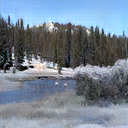} \\

\includegraphics[width=.19\linewidth]{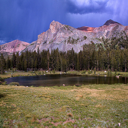} &
\includegraphics[width=.19\linewidth]{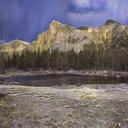} &
\includegraphics[width=.19\linewidth]{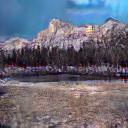} &
\includegraphics[width=.19\linewidth]{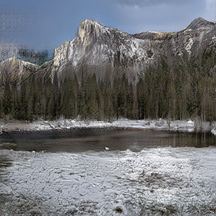} &
\includegraphics[width=.19\linewidth]{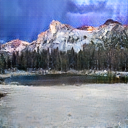} \\
\multicolumn{5}{c}{summer $\Rightarrow$ winter } \\
\includegraphics[width=.19\linewidth]{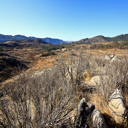} &
\includegraphics[width=.19\linewidth]{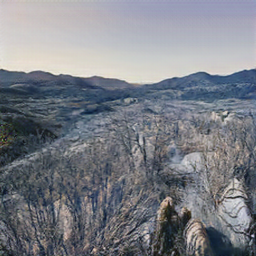} &
\includegraphics[width=.19\linewidth]{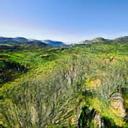} &
\includegraphics[width=.19\linewidth]{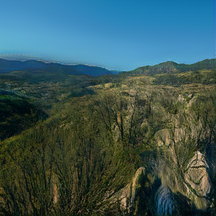} &
\includegraphics[width=.19\linewidth]{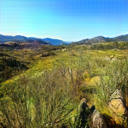} \\
\includegraphics[width=.19\linewidth]{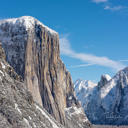} &
\includegraphics[width=.19\linewidth]{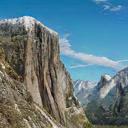} &
\includegraphics[width=.19\linewidth]{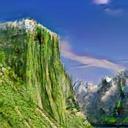} &
\includegraphics[width=.19\linewidth]{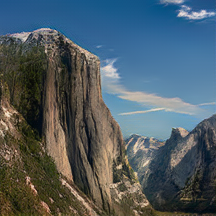} &
\includegraphics[width=.19\linewidth]{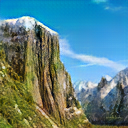} \\
\multicolumn{5}{c}{winter $\Rightarrow$ summer } \\
\end{tabular}
\caption{Example results of season transfer on summer and winter Yosemite photos from Flickr. For each penal, the first column shows some examples of testing input images, and the rest columns display the images ``translated'' by CycleGAN, UNIT, DRIT and our method, respectively.}
\label{fig:summer2winter}
\end{figure}

\begin{figure}[!htb]
\setlength{\tabcolsep}{1.5pt}
\centering
\begin{tabular}{cccc}
Input & ours & CycleGAN & Ground truth \\ 
\includegraphics[width=.21\linewidth]{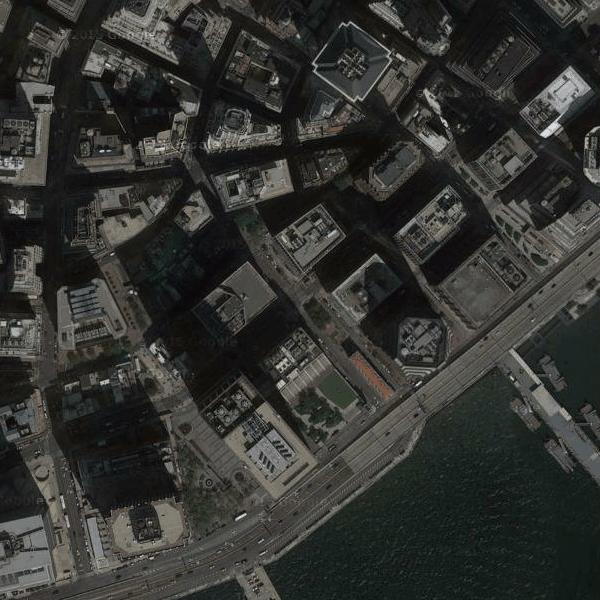} &
\includegraphics[width=.21\linewidth]{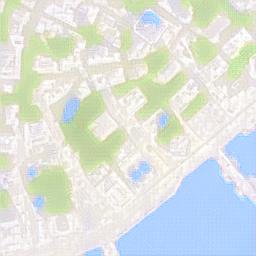} &
\includegraphics[width=.21\linewidth]{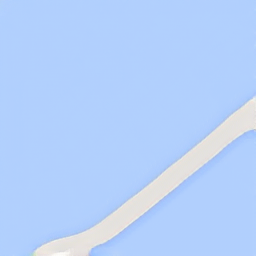} &
\includegraphics[width=.21\linewidth]{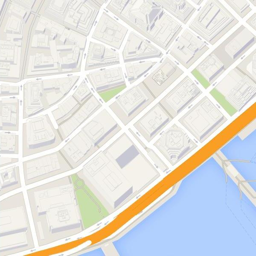} \\
\includegraphics[width=.21\linewidth]{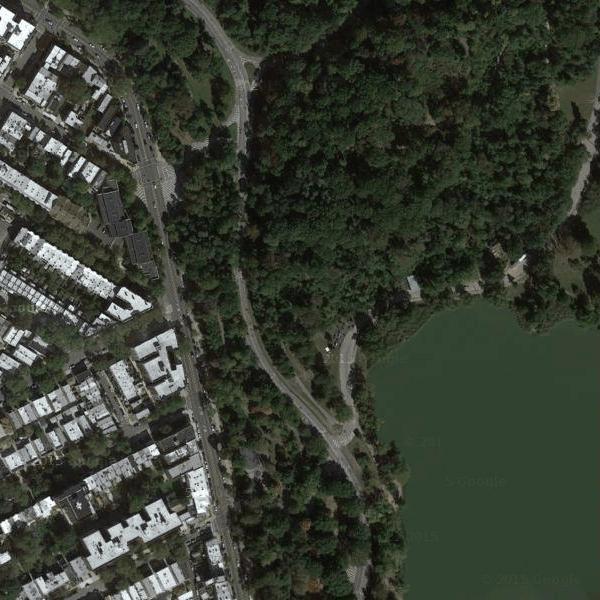} &
\includegraphics[width=.21\linewidth]{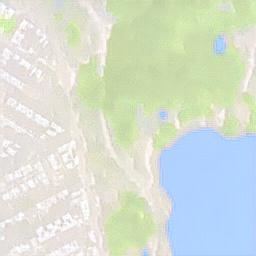} &
\includegraphics[width=.21\linewidth]{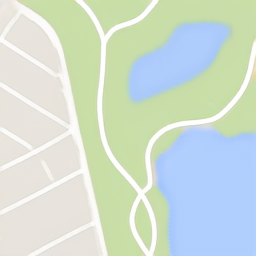} &
\includegraphics[width=.21\linewidth]{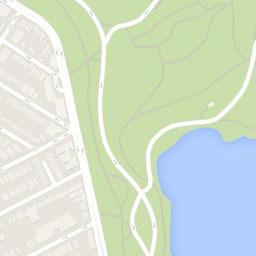} \\
\multicolumn{4}{c}{Aerial$\Rightarrow$Map}\\
Input & ours & CycleGAN & Ground truth \\ 
\includegraphics[width=.21\linewidth]{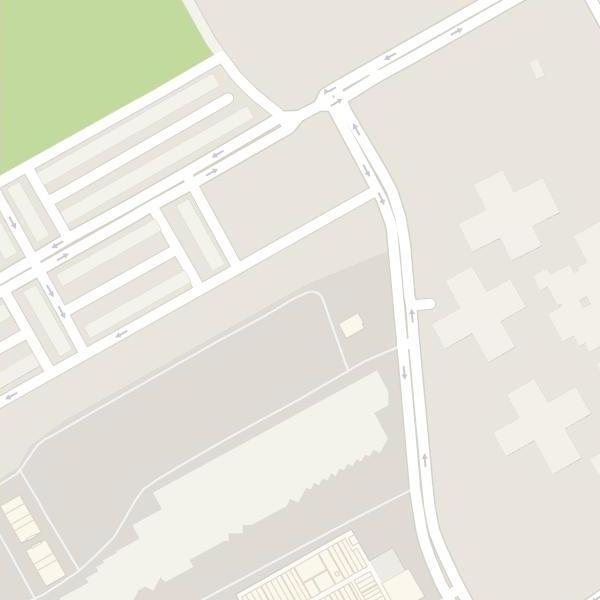} &
\includegraphics[width=.21\linewidth]{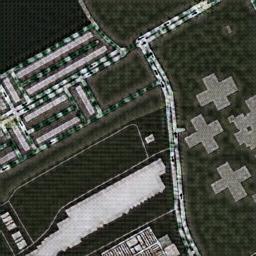} &
\includegraphics[width=.21\linewidth]{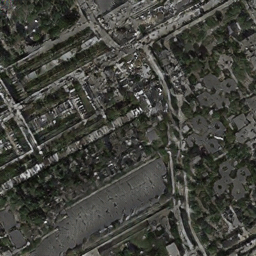} &
\includegraphics[width=.21\linewidth]{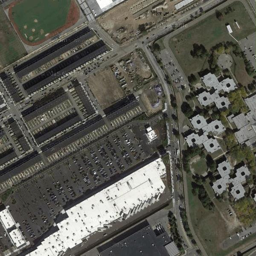}\\
\includegraphics[width=.21\linewidth]{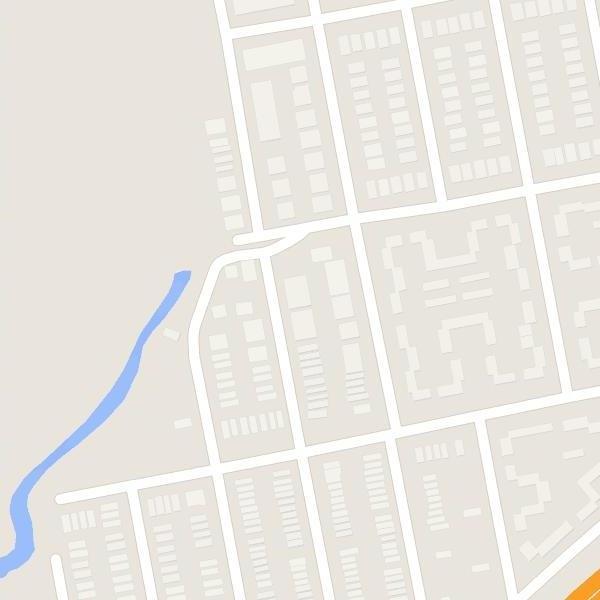} &
\includegraphics[width=.21\linewidth]{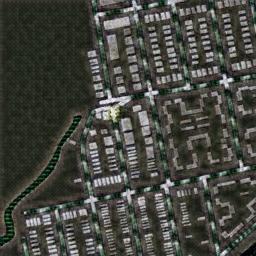} &
\includegraphics[width=.21\linewidth]{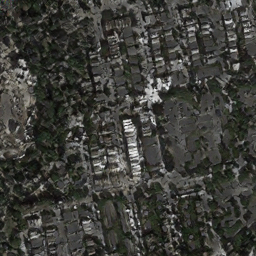} &
\includegraphics[width=.21\linewidth]{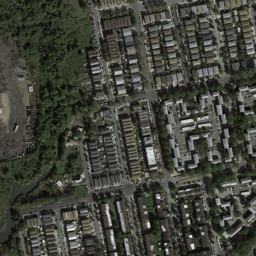}\\
\multicolumn{4}{c}{Map$\Rightarrow$Aerial}\\
\multicolumn{4}{c}{(a) Aerial$\Leftrightarrow$Map}\\

Input & ours & CycleGAN & Ground truth \\
\includegraphics[width=.21\linewidth]{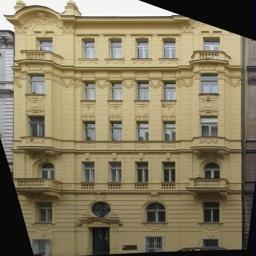} &
\includegraphics[width=.21\linewidth]{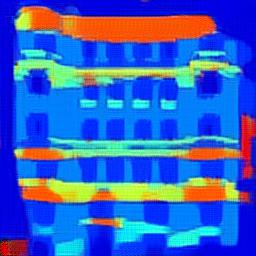} &
\includegraphics[width=.21\linewidth]{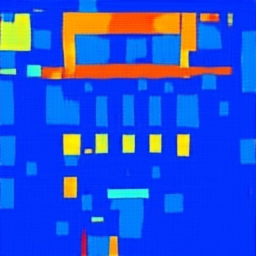} &
\includegraphics[width=.21\linewidth]{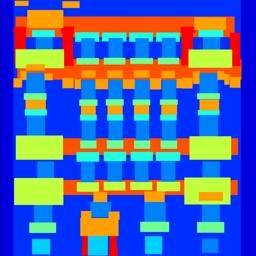} \\
\includegraphics[width=.21\linewidth]{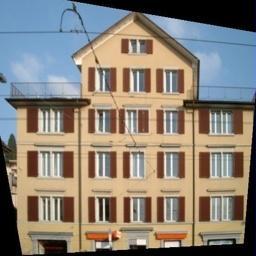} &
\includegraphics[width=.21\linewidth]{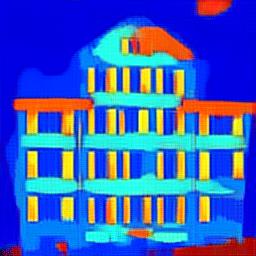} &
\includegraphics[width=.21\linewidth]{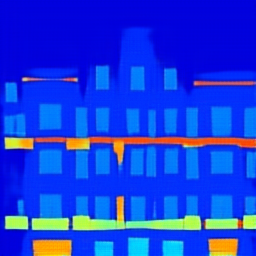} &
\includegraphics[width=.21\linewidth]{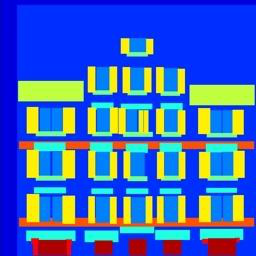}\\

\multicolumn{4}{c}{Facade$\Rightarrow$Label}\\
Input & ours & CycleGAN & Ground truth \\ 
\includegraphics[width=.21\linewidth]{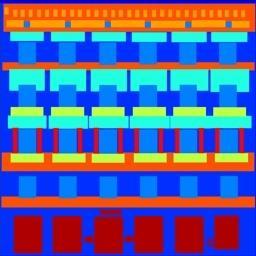} &
\includegraphics[width=.21\linewidth]{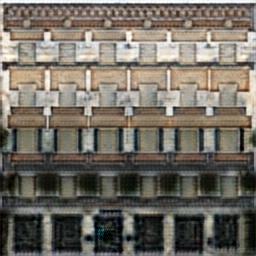} &
\includegraphics[width=.21\linewidth]{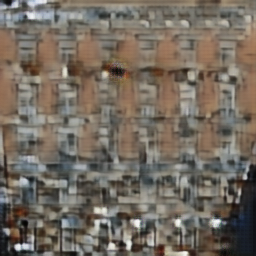} &
\includegraphics[width=.21\linewidth]{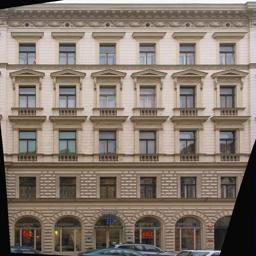}\\
\includegraphics[width=.21\linewidth]{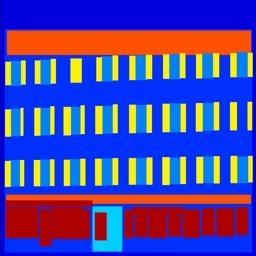} &
\includegraphics[width=.21\linewidth]{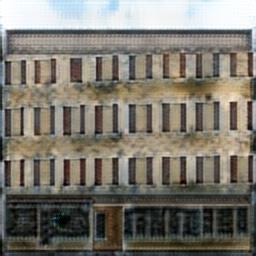} &
\includegraphics[width=.21\linewidth]{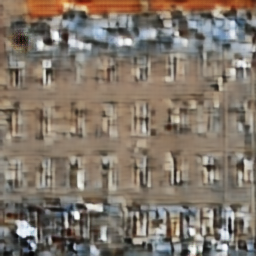} &
\includegraphics[width=.21\linewidth]{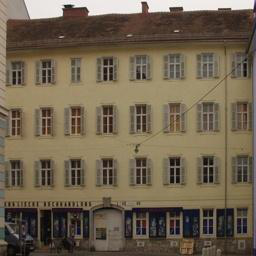}\\
\multicolumn{4}{c}{Label$\Rightarrow$Facade}\\
\multicolumn{4}{c}{(b) Facade$\Leftrightarrow$Label}
\end{tabular}
\caption{Qualitative results of unpaired image-to-image translations on datasets (a) aerial$\Leftrightarrow$ map (b) facade $\Leftrightarrow$ label.}
\label{fig:map}
\end{figure}

\subsubsection{Season transfer} We train our model on 854 winter photos and 1,273 summer photos of Yosemite that are used in \citet{zhu2017unpaired} for season transfer. Figure \ref{fig:summer2winter} shows some qualitative results and compares against three baseline methods, including CycleGAN, UNIT and DRIT \cite{lee2020drit++}. Our model obtains more realistic translation results compared with other baseline methods.

\begin{figure*}[ht!]
\setlength{\tabcolsep}{1.5pt}
\centering
\begin{tabular}{cccccc}
\multicolumn{1}{c}{} & {Input} & {Monet} & {Van Gogh} & {Cezanne} & {Ukiyo-e} \\
\hspace{0.5mm}\rotatebox{90}{\hspace{2.5mm}{example 1}}&
\includegraphics[width=.18\linewidth, height=.11\linewidth]{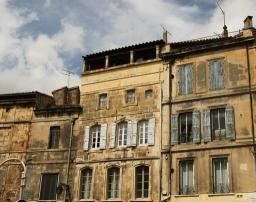} &
\includegraphics[width=.18\linewidth, height=.11\linewidth]{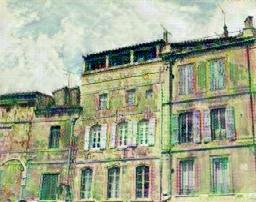}&
\includegraphics[width=.18\linewidth, height=.11\linewidth]{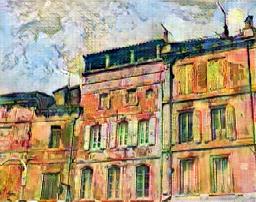}&
\includegraphics[width=.18\linewidth, height=.11\linewidth]{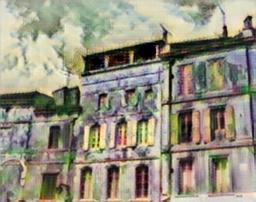}&
\includegraphics[width=.18\linewidth, height=.11\linewidth]{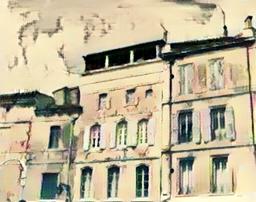}\\
\hspace{0.5mm}\rotatebox{90}{\hspace{2.5mm}{example 2}} &
\includegraphics[width=.18\linewidth, height=.11\linewidth]{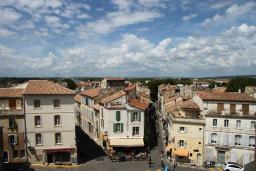} &
\includegraphics[width=.18\linewidth, height=.11\linewidth]{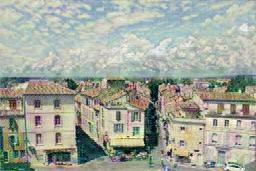} &
\includegraphics[width=.18\linewidth, height=.11\linewidth]{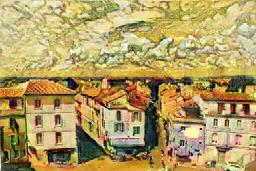} &
\includegraphics[width=.18\linewidth, height=.11\linewidth]{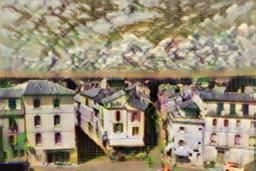} &
\includegraphics[width=.18\linewidth, height=.11\linewidth]{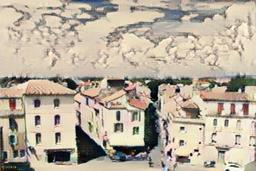}\\
\hspace{0.5mm}\rotatebox{90}{\hspace{2.5mm}{example 3}} &
\includegraphics[width=.18\linewidth, height=.11\linewidth]{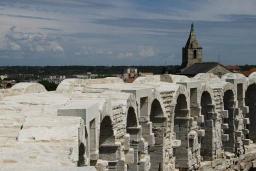} &
\includegraphics[width=.18\linewidth, height=.11\linewidth]{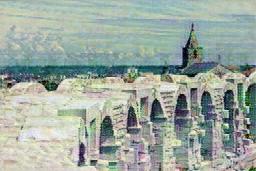} &
\includegraphics[width=.18\linewidth, height=.11\linewidth]{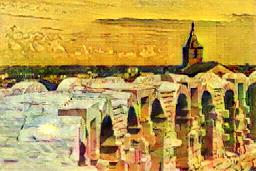} &
\includegraphics[width=.18\linewidth, height=.11\linewidth]{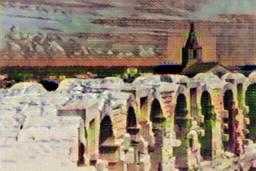} &
\includegraphics[width=.18\linewidth, height=.11\linewidth]{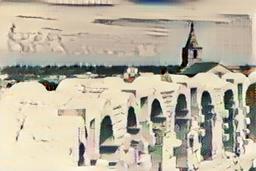}\\
\end{tabular}
\caption{Collection style transfer from photo realistic images to artistic styles.}
\label{fig:painting}
\end{figure*}

\begin{table}[ht!]
\begin{center}
\centering

\begin{tabular}{c|c|cc}
\hline
Dataset & Model & {$\uparrow$ L$\Rightarrow$R} & {$\uparrow$ R$\Rightarrow$L} \\ \hline
\multirow{4}{*}{Aerial$\Leftrightarrow$Map} & CycleGAN   & 21.59  & 12.67  \\ 
&AlignFlow(mle) & 19.47 & 13.60 \\	                                                             
&AlignFlow(adv) & 20.16& \textbf{15.17}\\		                                                   
&Ours                &     \textbf{22.29}          &   14.50  \\ \hline 
\multirow{4}{*}{Facade$\Leftrightarrow$Label}&CycleGAN   & 6.68  & 7.61  \\ 
&AlignFlow(mle) & 6.47 & 8.26 \\	                                                             
&AlignFlow(adv) & 7.74& 11.74\\		                                                   
&Ours  &     \textbf{9.34} &     \textbf{11.93} \\ \hline         
\end{tabular}
\caption{Quantitative evaluation in terms of PSNR on datasets Aerial$\Leftrightarrow$Map and Facade$\Leftrightarrow$Label. (adv: adversarial learning; mle: maximum likelihood estimation)}
\label{tab:comp_i2i}
\end{center}
\end{table}

\begin{figure*}[ht!]
\setlength{\tabcolsep}{1.5pt}
\centering
\begin{tabular}{ccccc}
 Input & Gatys et al.(Image I) &  Gatys et al.(Image II) &  CycleGAN&  CycleCoopNets (ours) \\
\includegraphics[width=.18\linewidth]{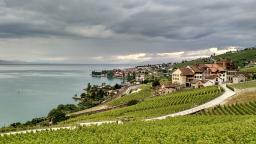} &
\includegraphics[width=.18\linewidth]{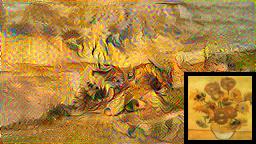} &
\includegraphics[width=.18\linewidth]{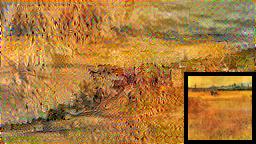} &
\includegraphics[width=.18\linewidth]{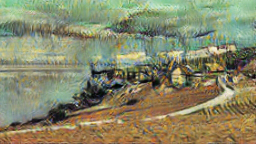} &
\includegraphics[width=.18\linewidth]{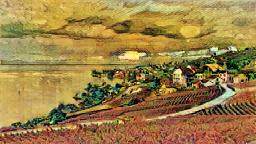} \\
\multicolumn{5}{c}{photo $\Rightarrow$ Van Gogh}\\ 
\includegraphics[width=.18\linewidth]{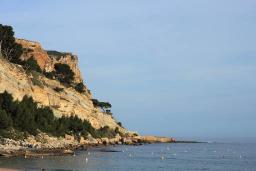} &
\includegraphics[width=.18\linewidth]{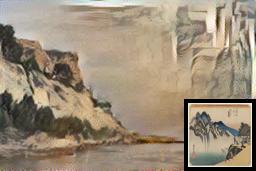} &
\includegraphics[width=.18\linewidth]{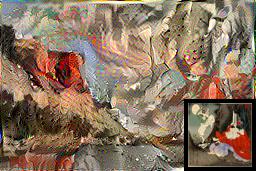} &
\includegraphics[width=.18\linewidth]{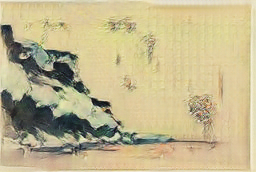} &
\includegraphics[width=.18\linewidth]{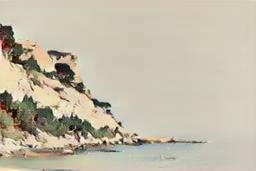} \\
\multicolumn{5}{c}{photo $\Rightarrow$ Ukiyo-e}\\ 
\end{tabular}
\caption{We compare our framework with style transfer method using neural network \cite{German2015a} on photo stylization. Each row represents one example, where the first column shows the input image, the second and the third columns show results
from \citet{German2015a} using two different representative artworks as style images, the fourth column displays the result of CycleGAN, and the last one is the result by our method CycleCoopNets.}
\label{fig:style_transfer}
\end{figure*}

\subsubsection{Translation between photo image and semantic label image}
We evaluate the proposed framework on two image-to-image translation datasets: (a) Aerial$\Leftrightarrow$Map \cite{isola2017image} and (b) Facade$\Leftrightarrow$label \cite{tylevcek2013spatial}. These two datasets provide one-to-one paired images that are originally used for supervised image-to-image translation in \citet{isola2017image}. In this experiment, we train our model on the datasets in an unsupervised manner, where the correspondence information between two image domains is omitted. We only use this correspondence information at the testing stage to compute the similarity between the generated images and the corresponding ground truths for quantitative evaluation. Table \ref{tab:comp_i2i} shows a comparison of our method and some baselines, which include CycleGAN and AlignFlow \cite{grover2020alignflow}. AlignFlow is a generative framework that uses normalizing flows for unsupervised image-to-image translation. We consider two types of training for the AlignFlow. One is based on maximum likelihood estimation, while the other is based on adversarial learning. We measure the similarity between two images via peak signal-to-noise ratio (PSNR), which is a suitable metric for evaluating datasets with one-to-one paring information. Figure \ref{fig:map} displays some qualitative results for both datasets. Our method shows comparable results with the baselines.

\subsubsection{Art style transfer} We evaluate our model on collection style transfer. We learn to translate landscape photographs into art paintings in the styles of Monet, Van Gogh, Cezanne and Ukiyo-e. The collections of landscape photographs are downloaded from Flickr and WikiArt and used in \citet{zhu2017unpaired}. 
We train a model between the photograph collection and each of the art collections to obtain the translator from photograph domain to painting domain. Figure \ref{fig:painting} displays some results. Each column represents one artistic style.

In Figure \ref{fig:style_transfer}, we compare our model with neural style transfer \cite{gatys2016image} and CycleGAN on photo stylization. Different from our method, \citet{gatys2016image} requires an image that specifies the target style to stylize an input photo image. Different rows show experiments with different target artistic styles. For each row, the input photo image is displayed in the first column, and we choose two representative artworks from the artistic collection as the style images for \citet{gatys2016image} and show their results in the second and third columns, respectively. CycleGAN and our method can stylize photos based on the style of the entire artistic collection, whose results are respectively shown in the last two columns. We find that nerual style transfer method \cite{gatys2016image} is difficult to generate meaningful results, while our method succeeds to produce meaningful ones that have a similar style to the target domain, which are  comparable with those obtained by CycleGAN.

\subsubsection{Time complexity} We highlight three points regarding the time complexity. (1) Although learning EBMs involves MCMC, each encoder-decoder $G$ in our framework serves to initialize the MCMC process, so that we only need a few steps of MCMC at each iteration. (2) Our MCMC method is the Langevin dynamics, which is a gradient-based algorithm, which means we only need to compute the gradient of the ConvNet-parameterized energy function with respect to the image. This can be efficiently accomplished by back-propagation due to the differentiability of the ConvNet. Other sampling methods or parametrization methods might not have such a convenience. (3) For implementation, we use TensorFlow as our framework and build the $l$-step MCMC process as a static computational graph that enables an efficient offline sampling. In all, the whole proposed framework is efficient and can be scaled up for large datasets with current PCs and GPUs. Taking the task of style transfer   on the VanGogh2photo dataset (roughly 6,200 training examples) as an example, for training images of size $256 \times 256$, our training time is roughly 0.80 seconds per iteration with 30 Langevin steps, while the CycleGAN takes 0.28 seconds per iteration. The execution time is recorded in a PC with an Intel i7-6700k CPU and a Titan Xp GPU.

\subsection{Unsupervised Image Sequence Translation}	 
We test our framework for image sequence translation. We use a U-Net structure as the temporal prediction model $R$, which takes as input a concatenation of two consecutive image frames in the past and predicts the next future frame. The U-Net structure follows the same design as the one used in \citet{isola2017image}. The EBM $p$ has a bottom-up ConvNet structure $f$ that consists of 3 layers of convolutions with numbers of channels $\{64, 128, 256\}$, filter sizes $\{5, 3, 3\}$, and subsampling factors $\{2, 2, 1\}$ at different layers, and one fully connected layer with 10 filters. Leaky ReLU layers are used between convolutional layers. The encoder-decoder $G$ is the same as the one we use in Section \ref{sec:image}. The step size for Langevin is 0.02. The number of Langevin steps is 15 for each EBM. We adopt Adam for optimization with a learning rate 0.0002. We set $\lambda_1=9$ and $\lambda_2=9$ in Eq.(\ref{eq:final_loss}). The mini-batch size is 1, and the number of parallel chains is 1. 
\begin{figure}
\centering	
\setlength{\tabcolsep}{.5pt}
\begin{tabular}{ccccccccc}
\hspace{0.5mm}\rotatebox{90}{\hspace{1.5mm}{\scriptsize Input}}
\includegraphics[width=.1\linewidth]{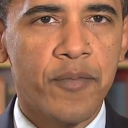} &
\includegraphics[width=.1\linewidth]{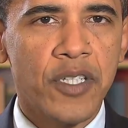} &
\includegraphics[width=.1\linewidth]{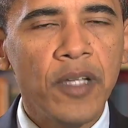} &
\includegraphics[width=.1\linewidth]{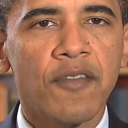} &
\includegraphics[width=.1\linewidth]{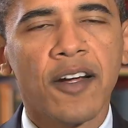} &
\includegraphics[width=.1\linewidth]{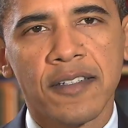} &
\includegraphics[width=.1\linewidth]{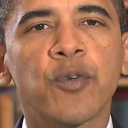} &
\includegraphics[width=.1\linewidth]{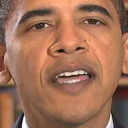} &
\includegraphics[width=.1\linewidth]{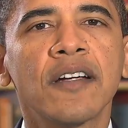} \\ 

\hspace{0.5mm}\rotatebox{90}{\hspace{1mm}{\scriptsize Output}}
\includegraphics[width=.1\linewidth]{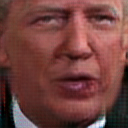} &
\includegraphics[width=.1\linewidth]{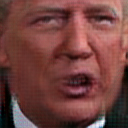} &
\includegraphics[width=.1\linewidth]{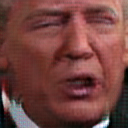} &
\includegraphics[width=.1\linewidth]{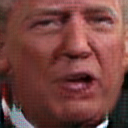} &
\includegraphics[width=.1\linewidth]{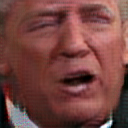} &
\includegraphics[width=.1\linewidth]{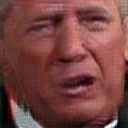} &
\includegraphics[width=.1\linewidth]{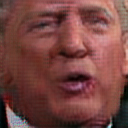} &
\includegraphics[width=.1\linewidth]{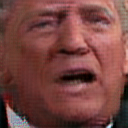} &
\includegraphics[width=.1\linewidth]{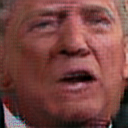}\\
\end{tabular}
{\footnotesize (a) Barack Obama to Donald Trump} \\ \vspace{1mm}

\begin{tabular}{ccccccccc}
\hspace{0.5mm}\rotatebox{90}{\hspace{1.5mm}{\scriptsize Input}}
\includegraphics[width=.1\linewidth]{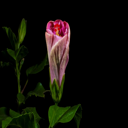} &
\includegraphics[width=.1\linewidth]{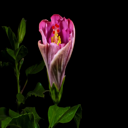} &
\includegraphics[width=.1\linewidth]{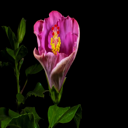} &
\includegraphics[width=.1\linewidth]{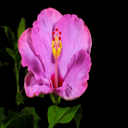} &
\includegraphics[width=.1\linewidth]{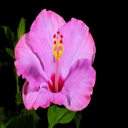} &
\includegraphics[width=.1\linewidth]{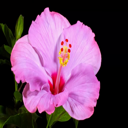} &
\includegraphics[width=.1\linewidth]{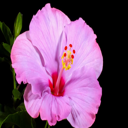} &
\includegraphics[width=.1\linewidth]{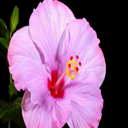} &
\includegraphics[width=.1\linewidth]{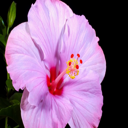} \\ \vspace{1mm}
\hspace{0.5mm}\rotatebox{90}{\hspace{1mm}{\scriptsize Output}}
\includegraphics[width=.1\linewidth]{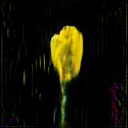} &
\includegraphics[width=.1\linewidth]{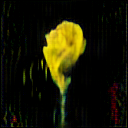} &
\includegraphics[width=.1\linewidth]{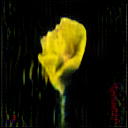} &
\includegraphics[width=.1\linewidth]{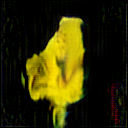} &
\includegraphics[width=.1\linewidth]{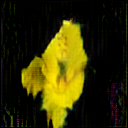} &
\includegraphics[width=.1\linewidth]{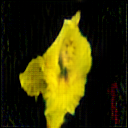} &
\includegraphics[width=.1\linewidth]{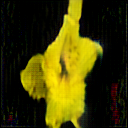} &
\includegraphics[width=.1\linewidth]{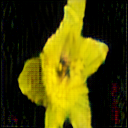} &
\includegraphics[width=.1\linewidth]{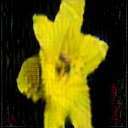}\\
\end{tabular}
{\footnotesize (b) violet flower to yellow flower}\\ \vspace{1mm}
\begin{tabular}{ccccccccc}
\hspace{0.5mm}\rotatebox{90}{\hspace{1.5mm}{\scriptsize Input}}
\includegraphics[width=.1\linewidth]{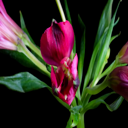} &
\includegraphics[width=.1\linewidth]{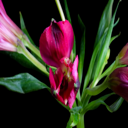} &
\includegraphics[width=.1\linewidth]{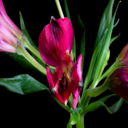} &
\includegraphics[width=.1\linewidth]{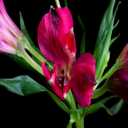} &
\includegraphics[width=.1\linewidth]{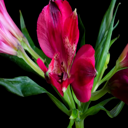} &
\includegraphics[width=.1\linewidth]{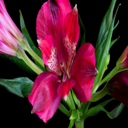} &
\includegraphics[width=.1\linewidth]{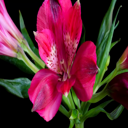} &
\includegraphics[width=.1\linewidth]{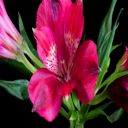} &
\includegraphics[width=.1\linewidth]{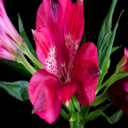}\\\vspace{1mm}
\hspace{0.5mm}\rotatebox{90}{\hspace{1mm}{\scriptsize Output}}
\includegraphics[width=.1\linewidth]{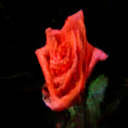} &
\includegraphics[width=.1\linewidth]{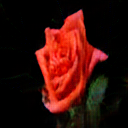} &
\includegraphics[width=.1\linewidth]{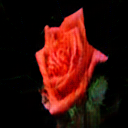} &
\includegraphics[width=.1\linewidth]{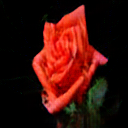} &
\includegraphics[width=.1\linewidth]{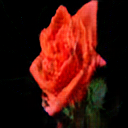} &
\includegraphics[width=.1\linewidth]{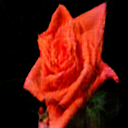} &
\includegraphics[width=.1\linewidth]{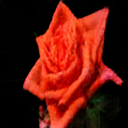} &
\includegraphics[width=.1\linewidth]{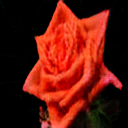} &
\includegraphics[width=.1\linewidth]{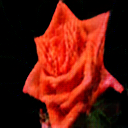}\\
\end{tabular}
{\footnotesize (c) purple flower to red flower}\\ 
\caption{Image sequence translation.  We translate (a) Barack Obama's facial motion to Donald Trump. (b) the blooming of a violet flower to a yellow flower. (c) the blooming of a purple flower to a red flower. For each case, the first row displays some image frames of the input sequence, and the second row shows the corresponding translated results.}
\label{fig:video}
\end{figure}

Figure \ref{fig:video} (a) shows an example of face-to-face translation from Barack Obama to Donald Trump. The first row shows some examples of image frames in the input image sequence, while the second row shows the corresponding image frames of the translated image sequence. For this experiment, the training sequences of faces are from \citet{bansal2018recycle}, in which the faces are extracted from publicly available videos of public figures, based on keypoints detected using the OpenPose library \cite{cao2017realtime}. The size of the image frame is $128 \times 128$ pixels. 
Figure \ref{fig:video} (b) and (c) show two examples of flower-to-flower translation. All training sequences are about blooming of different flowers. The image frames in each training sequence are ordered but all the sequences are neither synchronous nor aligned. The setting of the experiment is the same as the one of face-to-face translation. The results show that our framework can learn reasonable translation between two sequence domains without synchronous or aligned image frames. In particular, the translated sequences preserve the motion styles of the input sequences and only change their contents or appearances.

\section{Conclusion}

This paper studies unsupervised cross-domain translation problem based on a cooperative learning scheme. Our framework includes two cooperative networks, each of which consists of an energy-based model and a latent variable model to account for one domain distribution. We propose the alternating MCMC teaching algorithm to simultaneously train the two cooperative networks for maximum likelihood and cycle consistency. Experiments show that the proposed framework can be useful for different unsupervised cross-domain translation tasks. 

\section*{Acknowledgements}
The work is supported by NSF DMS-2015577, DARPA XAI N66001-17-2-4029, ARO W911NF1810296, ONR MURI N00014-16-1-2007, and XSEDE grant ASC180018. We gratefully acknowledge the support
of NVIDIA Corporation with the donation of the Titan Xp GPU
used for this research.

%
\bibliography{mybibfile}

\end{document}